\documentclass[10pt,twocolumn,letterpaper]{article}

\usepackage{iccv}
\usepackage{times}
\usepackage{epsfig}
\usepackage{graphicx}

\usepackage{amsmath}
\usepackage{amssymb}
\usepackage{multirow}
\usepackage{booktabs}
\usepackage{colortbl}
\definecolor{BrickRed}{rgb}{0.8,0.25,0.33}
\definecolor{BrightUbe}{rgb}{0.82,0.62,0.91}
\definecolor{Gray}{gray}{0.86}
\usepackage{graphics}

\usepackage[pagebackref=true,breaklinks=true,letterpaper=true,colorlinks,bookmarks=false]{hyperref}

\iccvfinalcopy 

\ificcvfinal\fi

\begin{document}

\title{Focus on Your Target: A Dual Teacher-Student Framework\\for Domain-adaptive Semantic Segmentation}

\author{
Xinyue Huo\textsuperscript{1,2}\quad Lingxi Xie\textsuperscript{2}\quad  Wengang Zhou\textsuperscript{1}\quad Houqiang Li\textsuperscript{1}\quad Qi Tian\textsuperscript{2}\\
\textsuperscript{1}University of Science and Technology of China\quad\textsuperscript{2}Huawei Inc.\\
\small\texttt{xinyueh@mail.ustc.edu.cn}\quad\small\texttt{198808xc@gmail.com}\\\small\texttt{\{zhwg,lihq\}@ustc.edu.cn}\quad\small\texttt{tian.qi1@huawei.com}
}


\maketitle
\ificcvfinal\thispagestyle{empty}\fi

\begin{abstract}
We study unsupervised domain adaptation (UDA) for semantic segmentation. Currently, a popular UDA framework lies in self-training which endows the model with two-fold abilities: (i) \textbf{learning} reliable semantics from the labeled images in the source domain, and (ii) \textbf{adapting} to the target domain via generating pseudo labels on the unlabeled images. We find that, by decreasing/increasing the proportion of training samples from the target domain, the `learning ability' is strengthened/weakened while the `adapting ability' goes in the opposite direction, implying a conflict between these two abilities, especially for a single model. To alleviate the issue, we propose a novel dual teacher-student (\textbf{DTS}) framework and equip it with a bidirectional learning strategy. By increasing the proportion of target-domain data, the second teacher-student model learns to `Focus on Your Target' while the first model is not affected. DTS is easily plugged into existing self-training approaches. In a standard UDA scenario (training on synthetic, labeled data and real, unlabeled data), DTS shows consistent gains over the baselines and sets new state-of-the-art results of 76.5\% and 75.1\% mIoUs on GTAv$\rightarrow$Cityscapes and SYNTHIA$\rightarrow$Cityscapes, respectively.
\end{abstract}

\section{Introduction}
\label{sec: introduction}

Deep neural networks~\cite{lecun2015deep} have been proven effective in learning from labeled, in-domain visual data, but their ability to adapt to unlabeled data or unknown domains is often not guaranteed. We delve into this topic by studying the unsupervised domain adaptation (UDA) problem for semantic segmentation, where densely (pixel-wise) labeled data are available for the source domain but only unlabeled images are provided for the target domain. The setting is useful in real-world applications where new domains emerge frequently but annotating them is a major burden.

\begin{figure}[!t]
\centering
\includegraphics[width=0.48\textwidth]{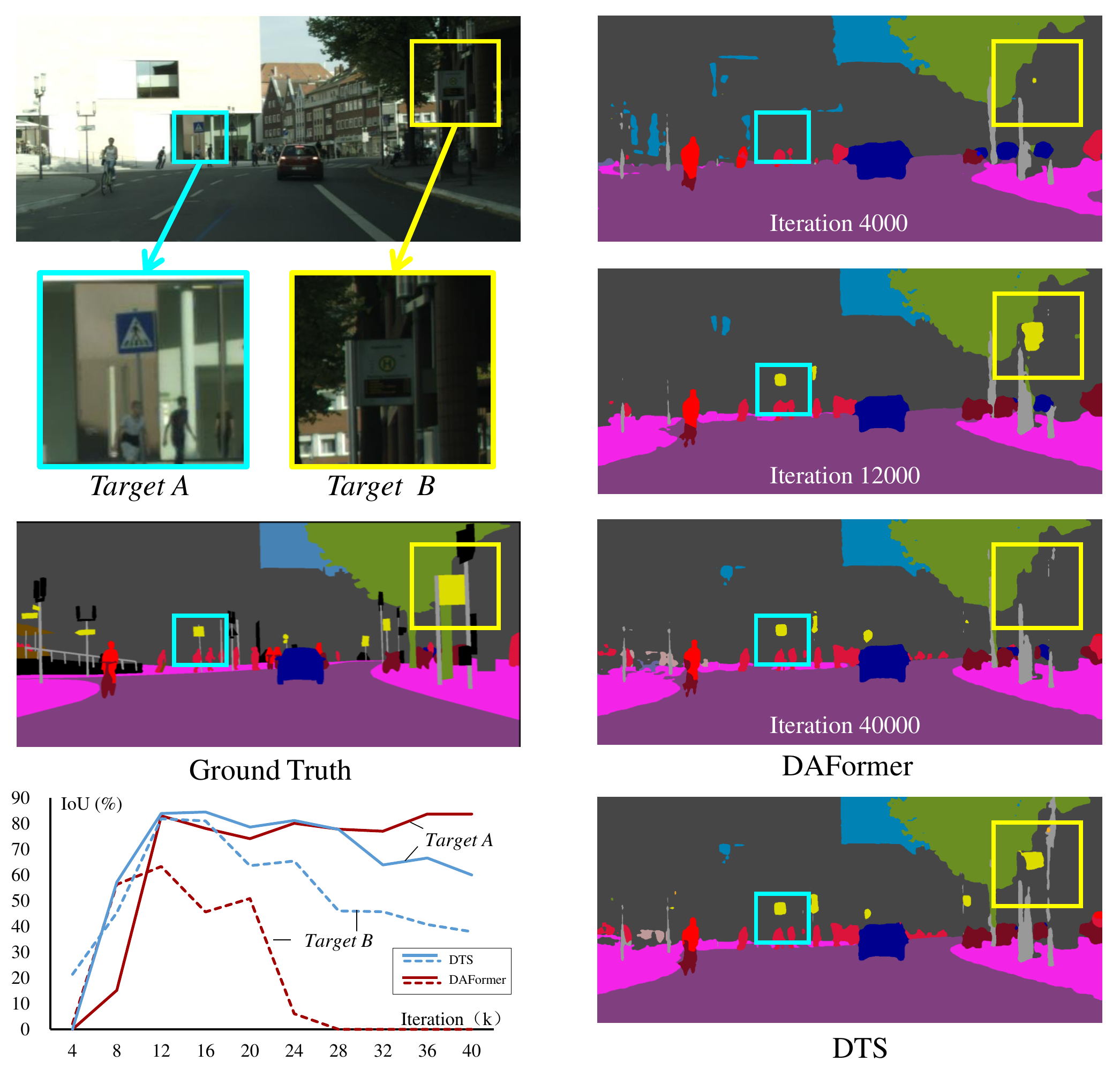}
\caption{Segmentation results on a test image from Cityscapes (the model is trained on GTAv). There are two \textsf{traffic signs} with different styles, where \textit{Target A} is similar to the training data in the source domain while \textit{Target B} is not. The baseline (\textcolor{BrickRed}{DAFormer}) gradually improves the accuracy on \textit{Target A} but achieves the best accuracy on \textit{Target B} in the midst (at $12\mathrm{K}$ iterations). Our method, \textcolor{cyan}{DTS}, achieves good accuracy for both targets and the trend is more stable throughout the training process (see the curves).}
\label{fig: motivation}
\end{figure}

The state-of-the-art UDA approaches for semantic segmentation are mostly built upon a self-training framework. It first learns semantics from labeled data in the source domain and then generates pseudo labels in the target domain, so that the model adapts to the target domain by fitting the pseudo labels. Conceptually, the two steps of self-training endow the model with two-fold abilities, namely, \textbf{learning} knowledge (semantics) from the source domain and \textbf{adapting} the knowledge to the target domain.

However, these two abilities can conflict with each other. Figure~\ref{fig: motivation} shows an example of two targets of the same class (\textsf{traffic sign}), where \textit{Target A} is close to the typical training samples in the source domain, while \textit{Target B} has a quite different appearance. Using the default setting of DAFormer~\cite{hoyer2022daformer} (a UDA baseline), the recognition accuracy of \textit{Target A} gradually increases while that of \textit{Target B} grows first but eventually drops to 0. Interestingly, if one increases the proportion of training samples from the target domain, the recognition quality of \textit{Target B} will be better while the recognition accuracy of some unexpected classes (\textit{e.g.}, \textsf{sidewalk}) may drop significantly (see details in Appendix~\ref{appendix: conflict}). The conflict sets an obstacle to achieving higher UDA accuracy.

The above analysis prompts us to design a dual teacher-student (\textbf{DTS}) framework. The key is to train two individual models (with different network weights) so that the aforementioned conflict is alleviated. DTS can be built upon most existing self-training approaches in three steps: (1) making two copies of the original teacher-student model, (2) increasing the proportion of training data from the target domain in the second group, and (3) applying a bidirectional learning strategy so that two models supervise each other via pseudo labels. DTS allows the system to pursue a stronger \textbf{adapting} ability (with the second model) meanwhile the \textbf{learning} ability (guaranteed by the first model) is mostly unaffected.

We conduct semantic segmentation experiments with the standard UDA setting. Two synthetic datasets with dense labels, GTAv~\cite{richter2016playing} and SYNTHIA~\cite{ros2016synthia}, are used as the source domains, and a real dataset with only images, Cityscapes~\cite{cordts2016cityscapes}, is used as the target domain. We establish the DTS framework upon three recent self-training baselines, namely, DAFormer~\cite{hoyer2022daformer}, HRDA~\cite{hoyer2022hrda}, and MIC~\cite{hoyer2022mic}. DTS improves the segmentation accuracy in every single trial, demonstrating its generalized ability across baselines and datasets. In particular, when integrated with MIC, the prior state-of-the-art, DTS reports $76.5\%$ and $67.8\%$ mIoUs on GTAv$\rightarrow$Cityscapes and SYNTHIA$\rightarrow$Cityscapes, respectively, setting new records for both benchmarks.

\section{Related Work}

\textbf{Unsupervised domain adaptation (UDA)}, the task of transferring semantics from labeled source domains to unlabeled target domains, has been widely studied for many computer vision fields, \textit{e.g.}, classification~\cite{long2017deep,zellinger2017central,tzeng2015simultaneous}, object detection~\cite{hoffman2014lsda,inoue2018cross,tang2016large}, face recognition~\cite{kan2015bi,hong2017sspp,sohn2017unsupervised}, semantic segmentation~\cite{hoffman2016fcns,zhang2017curriculum}, \textit{etc}.

We briefly review two representative UDA approaches for semantic segmentation. \textbf{Adversarial learning}~\cite{toldo2020unsupervised,tsai2018learning} reduces the gap between the source and target domains by confusing a discriminator trained to judge which domain the extracted visual features come from. It is closely related to style transfer methods~\cite{hoffman2018cycada,wu2018dcan,gong2019dlow,yang2020fda} that synthesize data to simulate the target domain. \textbf{Self-training} instead generates pseudo labels for the unlabeled target images and fits the model accordingly. There is a specific class of self-training methods~\cite{tranheden2021dacs,gao2021dsp,zhou2022context} which further bridge the domain gap by creating mixed images that cover both domains~\cite{olsson2021classmix}, inheriting benefits from style transferring. Many follow-up works~\cite{hoyer2022daformer,hoyer2022hrda,hoyer2022mic}, including this paper, were based on the self-training framework. Besides UDA, other data-efficient learning scenarios such as semi-supervised~\cite{huo2021atso,rosenberg2005semi} and few-shot~\cite{li2019learning,mukherjee2020uncertainty} learning also made use of self-training. Generating high-quality pseudo labels is crucial for self-training. Researchers developed various strategies, such as entropy minimization~\cite{chen2019domain}, class-balanced training~\cite{li2022class}, consistency regularization~\cite{hoyer2022mic}, auxiliary task learning~\cite{huo2022domain,vu2019dada}, \textit{etc.}, for this purpose.


A fundamental concept in self-training is \textbf{teacher-student optimization}, a concept closely related to knowledge distillation~\cite{hinton2015distilling,gou2021knowledge,mirzadeh2020improved}. The methodology was first applied to train a (smaller) student model which acquires stronger abilities by mimicking a (larger) teacher model. To boost the effect of teacher-student optimization, efforts were made in generating high-quality pseudo labels with stronger teacher models~\cite{tarvainen2017mean,ke2019dual,li2020dual,ge2020mutual} and/or preventing error transmission from the teacher model to the student model~\cite{ke2020guided,zou2020pseudoseg,huo2021atso,chen2021semi}. There are also other improvements including using multiple student models~\cite{zhang2018deep}, designing a better loss function~\cite{ge2020mutual}, \textit{etc}.


\section{Our Approach}
\label{approach}

We investigate the task of unsupervised domain adaptation (UDA) for semantic segmentation, where a labeled source domain $\mathbb{S}$ and an unlabeled target domain $\mathbb{T}$ are available. We denote the source dataset as $\mathcal{D}^\mathbb{S}=\{(\mathbf{x}_n^\mathbb{S},\mathbf{y}_n^\mathbb{S})\}_{n=1}^{N_\mathbb{S}}$, where $\mathbf{x}_n^\mathbb{S}$ is an image and $\mathbf{y}_n^\mathbb{S}$ is the corresponding dense label map. The target dataset $\mathcal{D}^\mathbb{T}$ is another set of images without labels, denoted as $\{(\mathbf{x}_n^\mathbb{T})\}_{n=1}^{N_\mathbb{T}}$. The two domains share the same set of categories. The goal of UDA is to train a segmentation network $f(\mathbf{x};\boldsymbol{\theta})$ that eventually works well in the target domain.

We first introduce a self-training baseline where a single teacher-student framework is established (Section~\ref{approach:single}). Then, we reveal a conflict when the model is facilitated to focus on the target domain (Section~\ref{approach:conflict}), which drives us to propose a dual teacher-student framework towards better UDA performance (Section~\ref{approach:dual}).

\subsection{Single Teacher-Student Baseline}
\label{approach:single}

We start with a standard self-training (\textit{i.e.}, teacher-student) baseline which trains a teacher model and a student model denoted as $f^\mathrm{te}(\mathbf{x};\boldsymbol{\theta}^\mathrm{te})$ and $f^\mathrm{st}(\mathbf{x};\boldsymbol{\theta}^\mathrm{st})$, respectively. The teacher model is updated using the exponential moving average (EMA) strategy~\cite{tarvainen2017mean}:
\begin{equation}
\label{eqn:ema}
\boldsymbol{\theta}^\mathrm{te} \leftarrow
 \boldsymbol{\theta}^\mathrm{te} \times \lambda + \boldsymbol{\theta}^\mathrm{st} \times ( 1 -  \lambda ),
\end{equation}
where $\lambda$ is a constant coefficient which is often close to $1$. The student model is trained for segmentation in both the source and target domains. For the source images, semantic labels are available for supervised learning:
\begin{equation}
\label{eqn:source-seg-loss}
\mathcal{L}^\mathbb{S} = \mathcal{L}_\mathrm{CE}(f^\mathrm{st}(\mathbf{x}^\mathbb{S}),\mathbf{y}^\mathbb{S}),
\end{equation}
where $\mathcal{L}_\mathrm{CE}$ denotes the pixel-wise cross-entropy loss. For the target images, the teacher model is used to generate a pseudo label $\tilde{\mathbf{y}}^\mathbb{T}$ for $\mathbf{x}^\mathbb{T}$, and the pixel-wise loss function is computed in a similar way:
\begin{equation}
\label{eqn:target-seg-loss}
\mathcal{L}^\mathbb{T} = \mathcal{L}_\mathrm{CE}(f^\mathrm{st}(\mathbf{x}^\mathbb{T}),\tilde{\mathbf{y}}^\mathbb{T}).
\end{equation}
The overall loss is written as $\mathcal{L} = \mathcal{L}^\mathbb{S} + \mathcal{L}^\mathbb{T}$. In practice, a balancing mechanism is applied~\cite{tranheden2021dacs,hoyer2022daformer}: all pixel-wise loss terms computed on the target domain is multiplied by a factor of $\gamma$, which is the proportion of the pseudo-label confidence surpassing a given threshold, $\tau$.

Recently, a series of methods~\cite{tranheden2021dacs,hoyer2022daformer,hoyer2022hrda,hoyer2022mic} validated that introducing a mixed domain between the source and target domains is helpful for domain adaptation. A typical operation is named ClassMix~\cite{olsson2021classmix} which samples mixed images by computing
\begin{equation}
\label{eqn:image-mix-st}
\mathbf{x}_{i,j}^{\langle\mathbb{S},\mathbb{T}\rangle}=\mathcal{M}(\mathbf{x}_i^\mathbb{S},\mathbf{x}_j^\mathbb{T})=\mathbf{x}_i^\mathbb{S}\odot \mathbf{M}_i^\mathbb{S}+\mathbf{x}_j^\mathbb{T}\odot(\mathbf{1}-\mathbf{M}_i^\mathbb{S}),
\end{equation}
where $\langle\mathbb{S,T}\rangle$ denotes a mixed domain, $\odot$ is element-wise multiplication, and $\mathbf{M}^\mathbb{S}$ is a mask generated by randomly a few class-level masks from the $\mathbf{y}^\mathbb{S}$. Accordingly, the target domain in Eqn~\eqref{eqn:target-seg-loss} and the overall loss is replaced with the mixed domain. Note that pseudo labels are only computed for the target portion of the mixed images -- we refer the reader to~\cite{olsson2021classmix} for technical details.

\begin{figure*}[!t]
\centering
\includegraphics[width=1.0\textwidth]{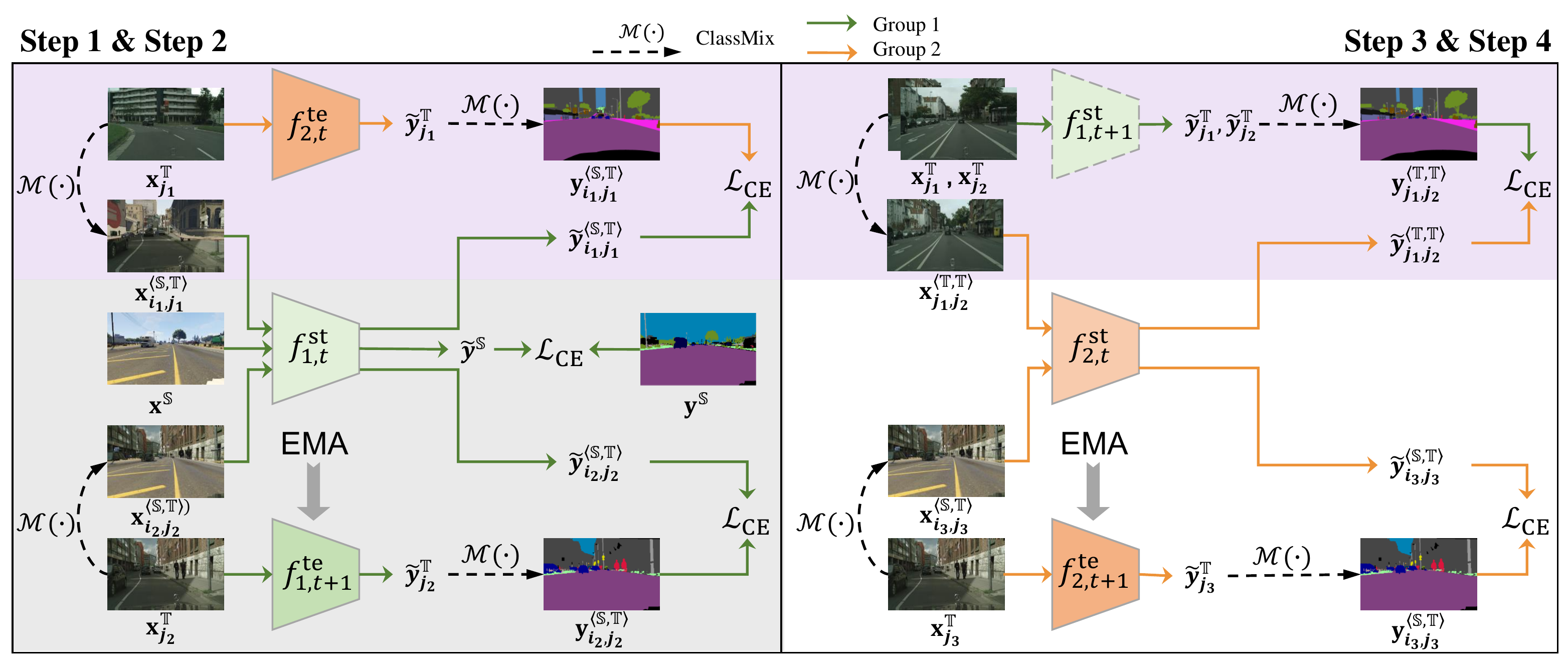}
\caption{An illustration of the proposed DTS framework, where we use \textit{Setting B} as an example. The left and right parts describe the optimization of the first and second group models, respectively. The bottom-left part (in \textcolor{Gray}{gray}) corresponds to a single teacher-student framework where ClassMix is used for domain mixing, and the top part (in \textcolor{BrightUbe}{purple}) shows the bidirectional learning strategy, \textit{i.e.}, models from the other group are used for pseudo label generation.}
\label{fig:framework}
\end{figure*} 

\subsection{Conflict between Learning and Adapting}
\label{approach:conflict}

Although domain mixing improves the accuracy of UDA segmentation on several benchmarks, we observe that the trained model is easily biased towards the source domain and thus ignores the target domain to some extent. A typical example is shown in Figure~\ref{fig: motivation}.

We conjecture that it is the `over-focus' on the source domain that prevents the model from achieving higher segmentation accuracy. To reveal this, we revisit the working mechanism of ClassMix. Compared to the original strategy that directly generates pseudo labels on the target images, ClassMix increases the proportion of data from the source domain (in a default setting of Eqn~\eqref{eqn:image-mix-st}, an average of half target-domain pixels are replaced with source-domain pixels). This brings two differences. On the one hand, the quality of pseudo labels is improved because part of them are inherited from the labels in the source domain; on the other hand, the focus on the target domain is weakened, increasing the risk of domain adaptation.

\begin{table}[!t]
\centering
\begin{tabular}{l|cc|c}
\toprule
Label & $\langle\mathbb{S},\mathbb{T}\rangle$ & $\langle\mathbb{T},\mathbb{T}\rangle$ & \textit{Supervised} \\
\midrule
w/ pseudo labels & \textbf{68.3} & 67.2 & -- \\
\quad (pseudo acc.) & 64.9 & 64.2 & -- \\
\midrule
\textit{w/ ground-truth} & \textit{74.4} & \textit{76.9} & \textbf{\textit{77.8}} \\
\bottomrule
\end{tabular}
\caption{UDA segmentation mIoU (\%) using DAFormer~\cite{hoyer2022daformer} on the GTAv$\rightarrow$Cityscapes benchmark with different options. An \textit{italic} number indicates a result that uses target-domain labels. The pseudo accuracy is computed on the training set of Cityscapes.}
\label{tab:mixture}
\end{table}

The above analysis prompts us to perform the \textit{`Focus on Your Target'} trail towards higher UDA accuracy. We replace the mixed domain of ClassMix, $\langle\mathbb{S},\mathbb{T}\rangle$, with another domain, $\langle\mathbb{T},\mathbb{T}\rangle$, in which the mixed samples are generated using two target images:
\begin{equation}
\label{eqn:image-mix-tt}
\mathbf{x}_{j_1,j_2}^{\langle\mathbb{T},\mathbb{T}\rangle}=\mathcal{M}(\mathbf{x}_{j_1}^\mathbb{T},\mathbf{x}_{j_2}^\mathbb{T})=\mathbf{x}_{j_1}^\mathbb{T}\odot \mathbf{M}_{j_1}^\mathbb{T}+\mathbf{x}_{j_2}^\mathbb{T}\odot(\mathbf{1}-\mathbf{M}_{j_1}^\mathbb{T}),
\end{equation}
and the pseudo label is adjusted accordingly.

In Table~\ref{tab:mixture}, we run DAFormer~\cite{hoyer2022daformer} with either $\langle\mathbb{S},\mathbb{T}\rangle$ or $\langle\mathbb{T},\mathbb{T}\rangle$ for UDA segmentation on GTAv$\rightarrow$Cityscapes. Compared to $\langle\mathbb{S},\mathbb{T}\rangle$, $\langle\mathbb{T},\mathbb{T}\rangle$ achieves an inferior mIoU of $67.2\%$ (a $1.3\%$ drop). But, interestingly, when ground-truth labels are used for the mixed images, the mIoU surpasses the original option by $2.5\%$, with merely a $0.9\%$ mIoU lower than the supervised baseline. In other words, \textit{`Focus on Your Target'} improves the upper bound, but the models cannot achieve the upper bound because the pseudo labels deteriorate: it is a side effect of decreasing the proportion of training data from the source domain.

In summary, a strong algorithm for UDA shall be equipped with two critical abilities: (i) \textbf{learning} reliable semantics from the source domain, and (ii) \textbf{adapting} the learned semantics to the target domain. However, in the above self-training framework where a single teacher-student model is used, there seems a conflict between these two abilities, \textit{e.g.}, increasing the proportion of target-domain data strengthens the `adapting ability' but weakens the `learning ability', setting an obstacle to better performance. In what follows, we offer a solution by training two teacher-student models simultaneously.

\subsection{Dual Teacher-Student Framework}
\label{approach:dual}

The proposed dual teacher-student (\textbf{DTS}) framework is illustrated in Figure~\ref{fig:framework}. We denote the models used in the two teacher-student groups as $f_1^\mathrm{te}(\mathbf{x};\boldsymbol{\theta}_1^\mathrm{te})$, $f_1^\mathrm{st}(\mathbf{x};\boldsymbol{\theta}_1^\mathrm{st})$ and $f_2^\mathrm{te}(\mathbf{x};\boldsymbol{\theta}_2^\mathrm{te})$, $f_2^\mathrm{st}(\mathbf{x};\boldsymbol{\theta}_2^\mathrm{st})$, respectively, where we use a subscript (1 or 2) to indicate the group ID. As described above, we expect the first teacher-student model to stick to \textbf{learning} reliable semantics from the source domain, while the second teacher-student model learns to \textit{`Focus on Your Target'} to enhance the \textbf{adapting} ability.

\begin{figure}[!t]
\centering
\includegraphics[width=0.46\textwidth]{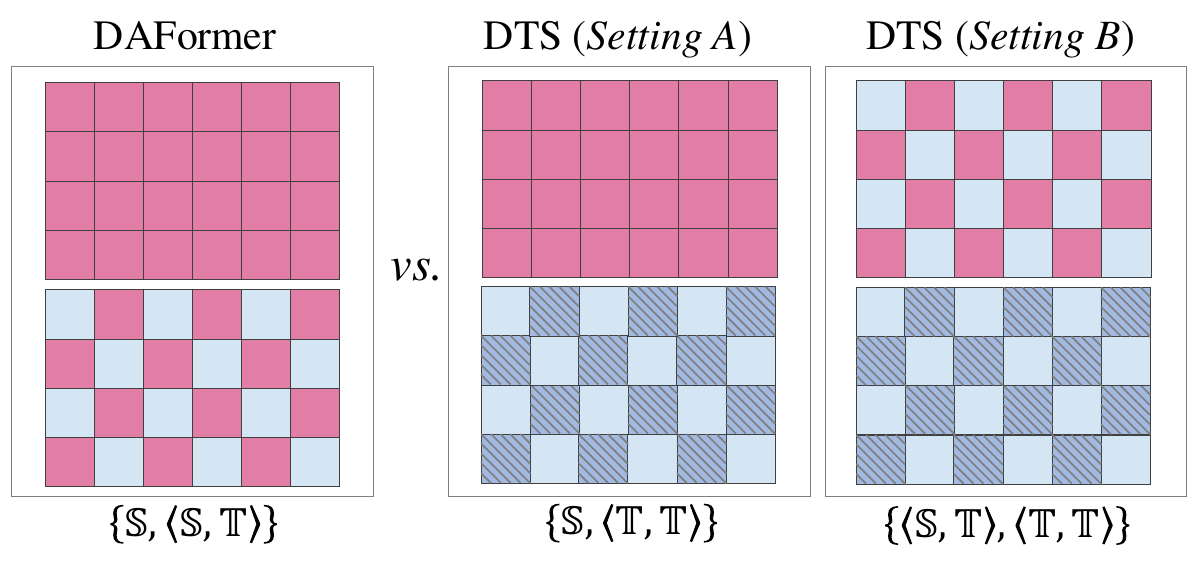}
\caption{An illustration of different combinations of training data. We use \textcolor{magenta}{magenta} to represent source-domain patches and two types of \textcolor{blue}{blue} to represent target-domain patches from different images. In practice, the images are not partitioned into regular grids, but into semantic regions according to the (real or pseudo) labels.}
\label{fig:combination}
\end{figure}

We achieve the goal by using different data proportions for the two groups. An illustration is provided in Figure~\ref{fig:combination}. The first group remains the same as in the baseline, \textit{i.e.}, half of the training samples are simply source images and another half are ClassMix-ed images covering the source and target domains. Using the previously defined notations, we denote the combination as $\{\mathbb{S},\langle\mathbb{S},\mathbb{T}\rangle\}$. To increase the proportion of the target-domain data, we first change the mixture of data from $\langle\mathbb{S},\mathbb{T}\rangle$ to $\langle\mathbb{T},\mathbb{T}\rangle$ (\textit{i.e.}, same as that used in the \textit{`Focus on Your Target'} trial), and then replace the pure source domain with $\langle\mathbb{S},\mathbb{T}\rangle$. That said, we create two settings for data combination, $\{\mathbb{S},\langle\mathbb{T},\mathbb{T}\rangle\}$ and $\{\langle\mathbb{S},\mathbb{T}\rangle,\langle\mathbb{T},\mathbb{T}\rangle\}$. We name them \textit{Setting A} and \textit{Setting B} of the second group throughout the remainder of this paper. Intuitively, the proportion of the target-domain data is increased from around $25\%$ in the first group (as well as the baselines) to around $50\%$/$75\%$ using \textit{Setting A}/\textit{Setting B} in the second group.

There are two teacher models and two student models to be optimized. We perform iterative optimization and propose a \textbf{bidirectional learning} strategy so that each teacher-student model propagates information to the other by contributing to the generation of pseudo labels. In the $t$-th iteration, we perform the following steps orderly:\\
\noindent\rule[0.2ex]{\linewidth}{0.2pt}\\
\textbf{(1) Updating $f_{1,t+1}^\mathrm{te}(\cdot)$} using the EMA of $f_{1,t}^\mathrm{st}(\cdot)$.\\
\textbf{(2) Updating $f_{1,t+1}^\mathrm{st}(\cdot)$.} This is done by generating a batch of images in the domain of $\langle\mathbb{S},\mathbb{T}\rangle$, feeding half of them into $f_{1,t+1}^\mathrm{te}(\cdot)$ and another half to $f_{2,t}^\mathrm{te}(\cdot)$ for generating pseudo labels. $f_{1,t+1}^\mathrm{st}(\cdot)$ is updated with gradient back-propagation using the pseudo labels and the true labels in $\mathbb{S}$.\\
\textbf{(3) Updating $f_{2,t+1}^\mathrm{te}(\cdot)$} using the EMA of $f_{2,t}^\mathrm{st}(\cdot)$.\\
\textbf{(4) Updating $f_{2,t+1}^\mathrm{st}(\cdot)$.} This is done by generating a batch of images in the domain of $\langle\mathbb{T},\mathbb{T}\rangle$ (and also $\langle\mathbb{S},\mathbb{T}\rangle$ if \textit{Setting B} is used), feeding half of them into $f_{2,t+1}^\mathrm{te}(\cdot)$ and another half to $f_{1,t+1}^\mathrm{st}(\cdot)$ for generating pseudo labels. $f_{2,t+1}^\mathrm{st}(\cdot)$ is updated with gradient back-propagation using the pseudo labels and the true labels in $\mathbb{S}$. The idea of using $f_{1,t+1}^\mathrm{st}(\cdot)$ (a model from the next iteration) for generating pseudo labels is known as `learning from future'~\cite{du2022learning} which improves the quality of pseudo labels for the current iteration.\\
\noindent\rule[0.2ex]{\linewidth}{0.2pt}
\textbf{Finally}, after a total of $T$ training iterations, $f_{2,T}^\mathrm{st}(\cdot)$ is chosen as the final model for inference.

Note that, if $f_{2,t}^\mathrm{te}(\cdot)$ is not used in Step 2 (\textit{i.e.}, all pseudo labels are generated using $f_{1,t+1}^\mathrm{te}(\cdot)$), the information is only propagated from the first teacher-student model to the second one, hence the optimization strategy is degenerated from bidirectional to unidirectional. The advantage of the bidirectional strategy, as well as other design choices, will be diagnosed in the experimental section.

\begin{table*}[!t]
\centering
\resizebox{\textwidth}{!}{
\setlength{\tabcolsep}{0.12cm}
\begin{tabular}{lcccccccccccccccccccc}
\toprule
\multicolumn{1}{l|}{\textbf{Method}} &
  \rotatebox{90}{Road}&
  \rotatebox{90}{Sidewalk} &
  \rotatebox{90}{Building} &
  \rotatebox{90}{Wall} &
 \rotatebox{90}{Fence} &
 \rotatebox{90}{Pole} &
  \rotatebox{90}{Light} &
  \rotatebox{90}{Sign} &
  \rotatebox{90}{Veg} &
  \rotatebox{90}{Terrain} &
 \rotatebox{90}{ Sky }&
  \rotatebox{90}{Person }&
  \rotatebox{90}{Rider} &
  \rotatebox{90}{Car }&
  \rotatebox{90}{Truck} &
  \rotatebox{90}{Bus} &
  \rotatebox{90}{Train} &
  \rotatebox{90}{Motor} &
  \multicolumn{1}{c|}{\rotatebox{90}{Bike}} &
  \textbf{mIoU} \\
\midrule
  
\multicolumn{1}{l|}{AdaptSegNet~\cite{tsai2018learning}}            & 86.5 & 36.0 & 79.9 & 23.4 & 23.3 & 23.9 & 35.2 & 14.8 & 83.4 & 33.3 & 75.6 & 58.5 & 27.6 & 73.7 & 32.5 & 35.4 & 3.9 & 30.1 &\multicolumn{1}{c|} {28.1} & 42.4 \\  

\multicolumn{1}{l|}{PatchAligne~\cite{tsai2019domain}}            & 92.3 & 51.9 &82.1 & 29.2 & 25.1 & 24.5 & 33.8 & 33.0 & 82.4 & 32.8 &82.2 & 58.6 & 27.2 & 84.3 & 33.4& 46.3 & 2.2 & 29.5 &\multicolumn{1}{c|} {32.3} & 46.5 \\

\multicolumn{1}{l|}{FDA~\cite{yang2020fda}}            & 92.5 & 53.3 & 82.4 & 26.5 & 27.6 & 36.4 & 40.6 & 38.9 & 82.3 & 39.8 & 78.0 & 62.6 & 34.4 & 84.9 & 34.1 & 53.1 & 16.9 & 27.7 &\multicolumn{1}{c|} {46.4} & 50.5 \\


\multicolumn{1}{l|}{DACS~\cite{tranheden2021dacs}} & 89.9 & 39.7 & 87.9 &  39.7& 39.5 & 38.5 & 46.4 & 52.8 & 88.0 & 44.0 & 88.8 & 67.2 & 35.8 & 84.5 & 45.7 & 50.2 & 0.0 & 27.3 & \multicolumn{1}{c|}{34.0} & 52.1  \\

\multicolumn{1}{l|}{ProDA~\cite{zhang2021prototypical}}    & 87.8 & 56.0 & 79.7 & {46.3} & {44.8} & {45.6} & {53.5} & {53.5} & {88.6} & {45.2} & 82.1 & {70.7} & {39.2} & {88.8} & {45.5} & {50.4} & 1.0 & {48.9} &  \multicolumn{1}{c|}{56.4} & 57.5 \\

\multicolumn{1}{l|}{DAP~\cite{huo2022domain}}    & 94.5 & 63.1 & 89.1 & 29.8 & 47.5 & 50.4 & 56.7 & 58.7 & 89.5 & 50.2 & 87.0 & 73.6 & 38.6 & 91.3 & 50.2 & 52.9 & 0.0 & 50.2 &  \multicolumn{1}{c|}{63.5} & 59.8 \\ 

\multicolumn{1}{l|}{CPSL~\cite{li2022class}}  & 92.3 & 59.9 & 84.9 & 45.7 & 29.7 & 52.8 & 61.5 & 59.5 & 87.9 & 41.6 & 85.0 & 73.0 & 35.5 & 90.4 & 48.7 & 73.9 & 26.3 & 53.8 & \multicolumn{1}{c|}{53.9} & 60.8 \\

\midrule

\multicolumn{1}{l|}{DAFormer~\cite{hoyer2022daformer}}    & 95.7 &70.2 &89.4 &53.5& 48.1& 49.6 &55.8 &59.4& 89.9& 47.9 & 92.5 & 72.2 & 44.7 & 92.3 & 74.5 & 78.2  & 65.1 & 55.9 & \multicolumn{1}{c|}{61.8}  & 68.3\\

\multicolumn{1}{l|}{DAFormer+FST~\cite{du2022learning}}  & 95.3 &67.7 &89.3 &55.5& 47.1& 50.1& 57.2 &58.6& 89.9 &51.0 &92.9 &72.7& 46.3& 92.5 &78.0 &81.6 &74.4 &57.7& \multicolumn{1}{c|}{62.6}& 69.3\\

\multicolumn{1}{l|}{DAFormer+CAMix~\cite{zhou2022context}}  &96.0 &73.1 &89.5& 53.9& 50.8& 51.7 &58.7 &64.9 &90.0 &51.2 &92.2 &71.8 &44.0 &92.8 &78.7 &82.3 &70.9 &54.1 & \multicolumn{1}{c|}{64.3 }&70.0\\

\multicolumn{1}{l|}{\textbf{DAFormer+DTS (ours)}}    & 96.8 &  76.0 & 	90.0 & 	56.5 &  	50.1 &  	51.9  & 	57.4  & 65.1 	 & 90.5  & 	50.5 	 & 92.4  & 	73.5  & 	46.9  & 	93.0 	 & 80.4  & 	85.2  & 	74.0  & 	58.7  & 	\multicolumn{1}{c|}{63.5 } & 	71.2  \\ 

\midrule
\multicolumn{1}{l|}{HRDA~\cite{hoyer2022hrda}}    & 96.4 & 74.4 & 91.0 & \textbf{61.6} & 51.5 & 57.1 & 63.9 & 69.3 & 91.3 & 48.4 & 94.2 & 79.0 & 52.9 & 93.9 & 84.1 & 85.7 & 75.9 & 63.9 & \multicolumn{1}{c|}{67.5 } 
& 73.8 \\

\multicolumn{1}{l|}{\textbf{HRDA+DTS (ours)}}    & 96.7 & 	76.3  & 	91.3  & 	60.7  & 	55.9  & 	59.9  & 	67.3  & 	72.5  & 	91.8  & 	50.1  & 	\textbf{94.4 } & 	80.7  & 	56.8 &  	94.4 &  	86.0 &  	86.0  & 	73.9  & \textbf{65.7} &  	 \multicolumn{1}{c|}{68.4}
  & 	75.2 
\\

\midrule
\multicolumn{1}{l|}{MIC~\cite{hoyer2022mic}}    &  \textbf{97.4 } & 80.1  & 91.7  & 61.2  & 56.9  & 59.7  & 66.0 &  71.3  & 91.7  & \textbf{51.4}  & 94.3  & 79.8  & 56.1  & 94.6  & 85.4  & 90.3  & 80.4  & 64.5  &  \multicolumn{1}{c|}{68.5}  & 75.9\\

\multicolumn{1}{l|}{\textbf{MIC+DTS (ours)}}    & 97.0  & 	\textbf{80.4 }	 & \textbf{91.8 } & 	60.6  & 	\textbf{58.7}  & 	\textbf{61.7}  & 	\textbf{67.9}  & \textbf{	73.2}  & 	\textbf{92.0}  & 	45.4  & 	94.3  & 	\textbf{81.3}  & 	\textbf{58.7 }	 & \textbf{95.0} 	 & \textbf{87.9} 	 & \textbf{90.7}  & \textbf{	82.2} 	 & \textbf{65.7}  & 	 \multicolumn{1}{c|}{\textbf{69.0}} &  	\textbf{76.5} 
 \\
\bottomrule
\end{tabular}
}
\caption{UDA segmentation accuracy (mIoU, \%) on the GTAv$\rightarrow$Cityscapes benchmark. The highest number in each column is marked with \textbf{bold}. The results in the first group are obtained on a ResNet101 backbone and the others are obtained on a MiT-B5 backbone.}
\label{tab:GTA}
\end{table*}

\subsection{Technical Details}
\label{approach:details}

\noindent\textbf{Choosing the proper option for data combination.}\quad
We compute the averaged confidence of pseudo labels generated by the corresponding teacher models, $f_1^\mathrm{st}(\cdot)$ and $f_2^\mathrm{te}(\cdot)$. Following~\cite{tranheden2021dacs,hoyer2022daformer}, $\gamma$ is defined as the proportion of pixels with the max class score surpassing a given threshold, $\tau$. For each training image $\mathbf{x}$, we use $\gamma_1^\mathrm{st}(\mathbf{x})$ and $\gamma_2^\mathrm{te}(\mathbf{x})$ to denote the average $\gamma$ values over all pixels when the teacher models are $f_1^\mathrm{st}(\cdot)$ and $f_2^\mathrm{te}(\cdot)$, respectively. Then, we estimate the chance that $f_2^\mathrm{te}(\cdot)$ generates better pseudo labels than $f_1^\mathrm{st}(\cdot)$:
\begin{equation}
\mathrm{Prob}=\mathbb{E}_{\mathbf{x}\sim\mathcal{D}^\mathbb{T}}\left[\mathbb{I}({\gamma}_2^\mathrm{te}(\mathbf{x})>{\gamma}_1^\mathrm{st}(\mathbf{x}))\right],
\end{equation}
where $\mathbb{I}(\cdot)$ is the indicator function that outputs $1$ if the statement is true and $0$ otherwise. Note that the estimation is performed individually for two training procedures using \textit{Setting A} and \textit{Setting B}. Finally, we choose the setting with a higher chance that $f_2^\mathrm{te}(\cdot)$ is more confident because $f_2^\mathrm{st}(\cdot)$ is eventually used for inference. We shall see in Section~\ref{experiments:diagnosis} that this criterion always leads to the better choice.

\noindent\textbf{Generating the mask.}\quad
In the domain mixing operation, we compute the mask $\mathbf{M}$ by choosing the class label of one image. If the $\langle\mathbb{S},\mathbb{T}\rangle$ is used, the source image with ground-truth is chosen; if $\langle\mathbb{T},\mathbb{T}\rangle$ is used, the image with a larger average value of $\gamma$ is chosen.

\noindent\textbf{Generating pseudo labels.}\quad
The details are slightly different between \textit{Setting A} and \textit{Setting B}. When the batch size is $k$, the number of target images used for the second group in \textit{Setting A} and \textit{Setting B} are $2k$ and $3k$, respectively. For \textit{Setting A}, $k$ pseudo labels used are generated by $f_{1,t+1}^\mathrm{st}(\cdot)$ and other $k$ by $f_{2,t+1}^\mathrm{te}(\cdot)$. For \textit{Setting B}, the $2k$ pseudo labels used in $\langle\mathbb{T},\mathbb{T}\rangle$ are generated by $f_{1,t+1}^\mathrm{st}(\cdot)$ and the $k$ labels used in $\langle\mathbb{S},\mathbb{T}\rangle$ are generated by $f_{2,t+1}^\mathrm{te}(\cdot)$.

\section{Experiments}
\label{experiments}

\subsection{Datasets and Settings}

\noindent\textbf{Datasets.}\quad
We train the models on a labeled source domain and an unlabeled target domain and test them on the target domain. We follow the convention to set the source domain to be a synthesized dataset (either GTAv~\cite{richter2016playing} which contains $24\rm{,}966$ densely labeled training images at a resolution of $1914\times1052$ or SYNTHIA~\cite{ros2016synthia} which contains $9\rm{,}400$ densely labeled training images at a resolution of $1280\times760$), and the target dataset to be a real dataset (Cityscapes~\cite{cordts2016cityscapes} where only the finely-annotated subset is used, containing $2\rm{,}975$ training images and $500$ validation images at a resolution of $2048\times1024$). Following the convention to report the mean IoU (mIoU) over $19$ classes when GTAv is used as the source domain, and $13$ or $16$ classes when SYNTHIA is the source domain.

\noindent\textbf{Training details.}\quad
Our framework is applied to different self-training frameworks. Specifically, we follow the DAFormer series~\cite{hoyer2022daformer,hoyer2022hrda,hoyer2022mic} where the encoder is a MiT-B5 model pre-trained on ImageNet-1K~\cite{deng2009imagenet} and the decoder involves a context-aware feature fusion module. We train the framework on a single NVIDIA Tesla-V100 GPU for $40\mathrm{K}$ iterations with a batch size of $2$ for both groups.  We use an AdamW optimizer~\cite{loshchilov2017decoupled} with learning rates of $6\times10^{-5}$ for the encoder and $6\times10^{-4}$ for the decoder, a weight decay of $0.01$, and linear warm-up of the learning rate for the first $1.5\mathrm{K}$ iterations. The input image size is always $512\times 512$. During the training phase, we apply random rescaling and cropping to the source and target data, as well as color jittering and Gaussian blur as strong augmentations for the mixed data. The EMA coefficient, $\lambda$, is set to be $0.999$. The pseudo-label confidence threshold, $\tau$, is $0.968$ as in~\cite{tranheden2021dacs}.

\begin{table*}[!t]
\centering
\resizebox{\textwidth}{!}{
\setlength{\tabcolsep}{0.16cm}
\begin{tabular}{lcccccccccccccccccc}
\toprule
\multicolumn{1}{l|}{Method} &
 \rotatebox{90}{Road}&
  \rotatebox{90}{Sidewalk} &
  \rotatebox{90}{Building} &
  \rotatebox{90}{Wall*} &
 \rotatebox{90}{Fence*} &
 \rotatebox{90}{Pole*} &
  \rotatebox{90}{Light} &
  \rotatebox{90}{Sign} &
  \rotatebox{90}{Veg} &
 \rotatebox{90}{ Sky }&
  \rotatebox{90}{Person }&
  \rotatebox{90}{Rider} &
  \rotatebox{90}{Car }&
  \rotatebox{90}{Bus} &  
  \rotatebox{90}{Motor} &
  \multicolumn{1}{c|}{\rotatebox{90}{Bike} }&
  \multicolumn{1}{l|}{mIoU} &
  mIoU* \\
\midrule
\multicolumn{1}{l|}{PatchAlign~\cite{tsai2019domain}}            & 82.4& 38.0 & 78.6 & 8.7 & 0.6 & 26.0 & 3.9 & 11.1 & 75.5 & 84.6 & 53.5 & 21.6 & 71.4 & 32.6 & 19.3 &31.7 & \multicolumn{1}{|c|}{40.0} & 46.5 \\    
  
\multicolumn{1}{l|}{AdaptSegNet~\cite{tsai2018learning}}            & 84.3 & 42.7 & 77.5 & -- & -- & -- & 4.7 & 7.0 & 77.9 & 82.5 & 54.3 & 21.0 & 72.3 & 32.2 & 18.9 & 32.3 & \multicolumn{1}{|c|}{--} & 46.7 \\     
  
\multicolumn{1}{l|}{FDA~\cite{yang2020fda}}            &  79.3 & 35.0 & 73.2 & -- & -- & -- & 19.9 & 24.0 & 61.7 & 82.6 & 61.4 & 31.1 & 83.9 & 40.8 & 38.4 & 51.1 & \multicolumn{1}{|c|}{--} & 52.5 \\


\multicolumn{1}{l|}{DACS~\cite{tranheden2021dacs}}            & 80.6 & 25.1 & 81.9 & 21.5 & 2.9 & 37.2 & 22.7 & 24.0 & 83.7 & 90.8 &67.6 & 38.3 & 82.9 & 38.9 & 28.5 & 47.6 & \multicolumn{1}{|c|}{48.3} & 54.8 \\ 

\multicolumn{1}{l|}{ProDA~\cite{zhang2021prototypical}} &   87.8 & 45.7 & 84.6 & {37.1} & 0.6 & 44.0 & {54.6} & {37.0} & {88.1} & 84.4 & 74.2 & 24.3 & 88.2 & {51.1} & {40.5} & 45.6 & 
\multicolumn{1}{|c|}{55.5} & 62.0 \\

\multicolumn{1}{l|}{DAP~\cite{huo2022domain}} &  84.2& 46.5 & 82.5 & 35.1 & 0.2 & 46.7 & 53.6 & 45.7 &  \textbf{89.3} & 87.5 & 75.7 & 34.6 &  \textbf{91.7} & 73.5 & 49.4 & 60.5 & 
\multicolumn{1}{|c|}{59.8}& 64.3 \\

\multicolumn{1}{l|}{CPSL~\cite{li2022class}}    & 87.2  & 43.9  & 85.5  & 33.6  & 0.3 &  47.7  & 57.4  & 37.2  & 87.8  & 88.5  & 79.0  & 32.0  & 90.6 &  49.4 &  50.8 &  59.8 &  \multicolumn{1}{|c|}{57.9}  & 65.3 \\

\midrule

\multicolumn{1}{l|}{DAFormer~\cite{hoyer2022daformer}}    &  84.5  & 	40.7  & 	88.4  & 	41.5  & 	6.5	  & 50.0   & 	55.0  & 	54.6  & 	86.0  & 	89.8  & 	73.2  & 	48.2	  & 87.2  & 	53.2  & 	53.9	  & 61.7	  & \multicolumn{1}{|c|}{60.9}	  & 67.4
 \\
\multicolumn{1}{l|}{DAFormer+FST~\cite{du2022learning}}  & 88.3  & 46.1 &  88.0  & 41.7  & 7.3  & 50.1  & 53.6  & 52.5  & 87.4  &  91.5  & 73.9 &  48.1 &  85.3 &   58.6 &   55.9 & 63.4 &  \multicolumn{1}{|c|}{61.9} & 68.7\\

\multicolumn{1}{l|}{DAFormer+CAMix~\cite{zhou2022context}}  & 87.4 &  47.5  & 88.8 &--&  --& -- &   55.2 &  55.4 &  87.0 &  91.7 &  72.0  & 49.3  & 86.9  & 57.0  & 57.5  & 63.6  & \multicolumn{1}{|c|}{--} &  69.2\\

\multicolumn{1}{l|}{\textbf{DAFormer+DTS (ours)}}    & 88.5  & 	51.4  & 	88.0  & 	37.2  & 	4.8  & 	51.4  & 	59.4  & 	59.0 	 & 88.1 	 & 92.7  & 	71.3  & 	51.3  & 	89.2  & 	 \textbf{67.7 } & 	58.4  & 	62.8  & 	\multicolumn{1}{|c|}{63.8 } & 	71.4 
\\ 

\midrule
\multicolumn{1}{l|}{HRDA~\cite{hoyer2022hrda}}    &  85.2	 & 47.7	 & 88.8 &  \textbf{49.5} & 	4.8 & 	57.2 & 	65.7	 & 60.9 & 	85.3 & 92.9	 & 79.4	 & 52.8	& 89.0 & 	64.7 & 	63.9 & 	 \textbf{64.9} & 	\multicolumn{1}{|c|}{65.8 } & 72.4 \\

\multicolumn{1}{l|}{\textbf{HRDA+DTS (ours)}}    & 88.5  & 	53.1 &  	88.7  & 	44.0  & 	5.8  & 	59.2  & 	 \textbf{68.3}  & 	60.7  & 	87.7  & 	93.4  & 	79.5  & 	46.4  & 	91.3  & 	64.2  & 	66.8  & 	65.9  & 		\multicolumn{1}{|c|}{66.5}  & 	73.4 

\\

\midrule
\multicolumn{1}{l|}{MIC~\cite{hoyer2022mic}}    & 86.6	 & 50.5 &  \textbf{89.3} & 	47.9 & 	7.8	 & 59.4	 & 66.7 & 63.4	 & 87.1	 &  \textbf{94.6} & 	81.0 & 58.9 & 	90.1 & 	61.9 & 	67.1 & 	64.3 & 	\multicolumn{1}{|c|}{67.3} & 	74.0 
  \\

\multicolumn{1}{l|}{\textbf{MIC+DTS (ours)}}    &   \textbf{89.1} & 	 \textbf{54.9} &  	89.0 &  	39.1  & 	 \textbf{8.7 } & 	 \textbf{61.6 } & 	67.4  & 	 \textbf{64.3} & 	88.8 &   94.0 	 &  \textbf{82.2}  &  \textbf{	60.7} 	 & 89.6  & 	62.6 	 & \textbf{68.5}  &  \textbf{64.9}  & 	\multicolumn{1}{|c|}{ \textbf{67.8 }}&  \textbf{	75.1 }
 \\
 
\bottomrule
\end{tabular}}

\caption{UDA segmentation accuracy (mIoU, \%) on the SYNTHIA$\rightarrow$Cityscapes benchmark. The highest number in each column is marked with \textbf{bold}. The results in the first group are obtained on a ResNet101 backbone and the others are obtained on a MiT-B5 backbone. The mIoU and mIoU* columns denote the averaged IoU over $16$ and $13$ classes, where the latter does not contain the classes marked with *.}
\label{tab:SYNTHIA}
\end{table*}

\subsection{Comparison to the State-of-the-art Methods}
\label{sec: comparison with sota}

We compare DTS to state-of-the-art UDA approaches on the GTAv$\rightarrow$Cityscapes and SYNTHIA$\rightarrow$Cityscapes benchmarks. Results are summarized in Table~\ref{tab:GTA} and Table~\ref{tab:SYNTHIA}, respectively. We establish DTS upon three recently published works, namely, DAFormer~\cite{hoyer2022daformer}, HRDA~\cite{hoyer2022hrda}, and MIC~\cite{hoyer2022mic}.

DTS reports consistent gains in terms of segmentation accuracy beyond all three baselines. On GTAv$\rightarrow$Cityscapes and SYNTHIA$\rightarrow$Cityscapes ($13$ classes), the mIoU gains over DAFormer, HRDA, MIC are $2.9\%$, $1.4\%$, $0.6\%$ and $4.0\%$, $1.0\%$, $1.1\%$ respectively. The gain generally becomes smaller when the baseline is stronger, which is as expected. Specifically, when added to MIC, the previous best approach, DTS achieves new records on these benchmarks: a $76.5\%$ mIoU on GTAv$\rightarrow$Cityscapes and $67.8\%$ and $75.1\%$ mIoUs on SYNTHIA$\rightarrow$Cityscapes with $16$ and $13$ classes, respectively.

\noindent\textbf{Comparison to FST.}\quad
We compare DTS to another plug-in approach, FST~\cite{du2022learning}. On DAFormer, the gains brought by FST on the two benchmarks (SYNTHIA using $13$ classes) are $1.0\%$ and $1.3\%$, while the numbers for DTS are $2.9\%$ and $4.0\%$, respectively. That said, besides the contribution of FST that allows us to `learn from future', the higher accuracy also owes to the design of the DTS framework that alleviates the conflict of \textbf{learning} and \textbf{adapting}.

\noindent\textbf{Class-level analysis.}\quad
We investigate which classes are best improved by DTS. On GTAv$\rightarrow$Cityscapes, we detect two classes, \textsf{sign} and \textsf{truck}, on which the IoU gain on the three baselines is always higher than the mIoU gain. In Appendix~\ref{appendix: class frequency}, we show that these classes suffer larger variations between the source-domain and target-domain appearances, and hence the learned features are more difficult to transfer. They are largely benefited by the \textit{`Focus on Your Target'} ability of DTS.

\begin{figure*}[!t]
\centering
\includegraphics[width=1.0\textwidth]{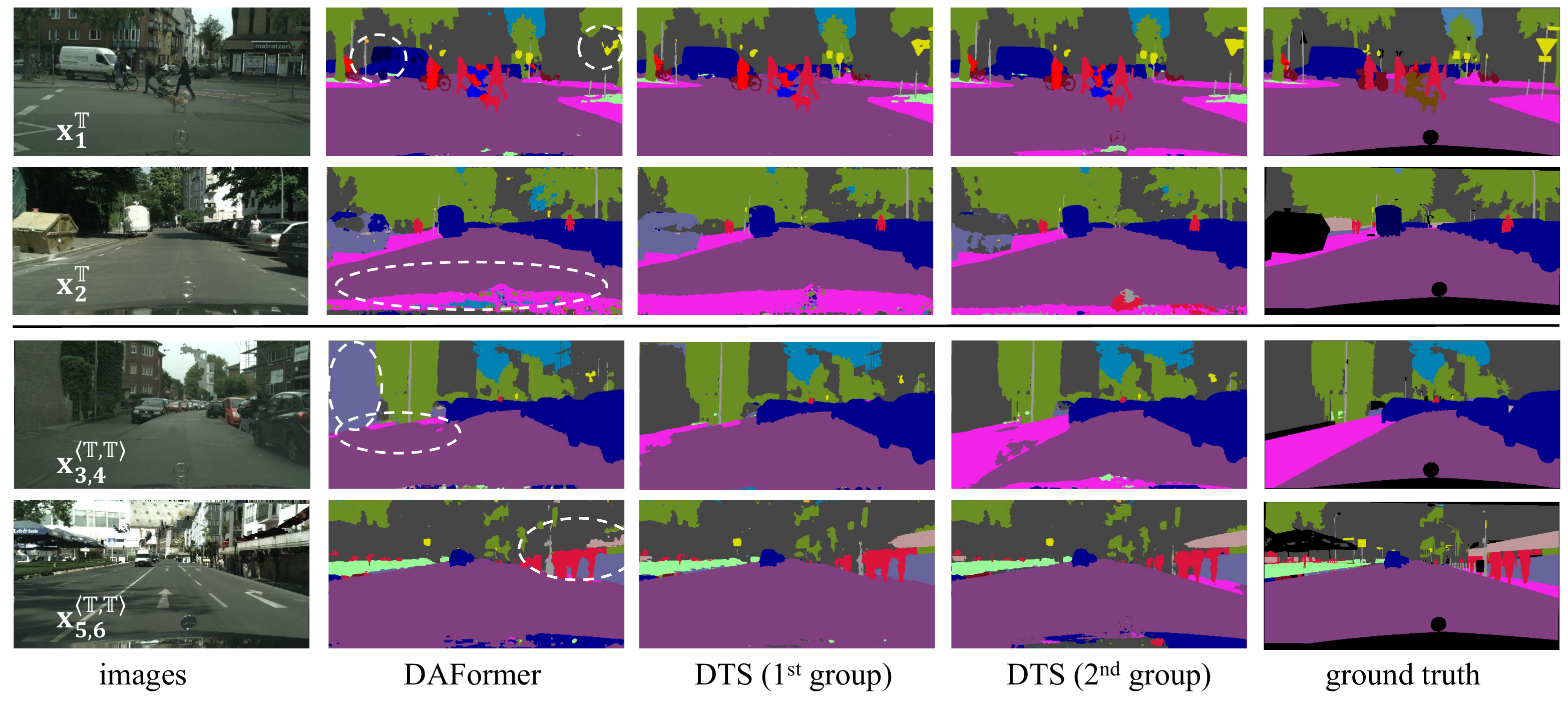}
\caption{Visualization of the segmentation results of DAFormer~\cite{hoyer2022daformer} and DTS. The top two rows display two real images and the bottom two rows display two synthesized images from $\langle\mathbb{T},\mathbb{T}\rangle$. In each group, the first example is from GTAv$\rightarrow$Cityscapes and the second one is from SYNTHIA$\rightarrow$Cityscapes. The white, dashed circles highlight the regions that are significantly improved by DTS.}
\label{fig:visualization}
\end{figure*}

\noindent\textbf{Visualization.}\quad
We visualize the segmentation results of DAFormer and DAFormer+DTS in Figure~\ref{fig:visualization}. For the real images, the advantage of DTS is significant on the classes that are easily confused with others (\textit{e.g.}, \textsf{sidewalk} which is visually similar to \textsf{road}) or suffer large domain shifts (\textit{e.g.}, \textsf{sign} or \textsf{truck}). 
For the mixed images, we find that the baseline model is easily confused by the unusual stitching and/or discontinuous textures because it has never seen such patterns in the source domain -- in such cases, the model of the second group is generally better and sometimes it can propagate the advantages to the model of the first group, revealing the advantage of bidirectional learning.

\subsection{Diagnostic Studies}
\label{experiments:diagnosis}

\begin{table}[!t]
\centering
\setlength{\tabcolsep}{1.2mm}
\begin{tabular}{l|ccccc}
\toprule
Setting         & \multicolumn{5}{c}{Choice and Accuracy} \\
\midrule
Dual Teacher-Student & & & \checkmark & \checkmark & \checkmark \\
\textit{`Focus on Your Target'} & & \checkmark & \checkmark &  &   \checkmark \\
Bidirectional Learning & & & & \checkmark & \checkmark \\
\midrule
GTA$\rightarrow$Cityscapes & 68.3 & 67.2 & 69.7 & 68.6 & \textbf{71.2} \\
SYNTHIA$\rightarrow$Cityscapes & 60.9 & 58.4 & 63.3 & 61.2 & \textbf{63.7} \\
\bottomrule
\end{tabular}
\caption{Ablative studies of DTS with respect to the UDA segmentation accuracy (mIoU, \%) on both benchmarks. All these experiments are based on DAFormer~\cite{hoyer2022daformer}. For SYNTHIA$\rightarrow$Cityscapes, the mIoUs are computed over $16$ classes.}
\label{tab:ablation}
\end{table}

\noindent\textbf{Ablative studies.}\quad
We first diagnose the proposed approach by ablating the contribution of three components, namely, applying the dual teacher-student (DTS) framework, increasing the proportion of data from the target domain (\textit{i.e.}, \textit{`Focus on Your Target'}), and adopting the bidirectional learning strategy. Results are summarized in Table~\ref{tab:ablation}. Note that some combinations are not valid, \textit{e.g.}, solely applying DTS (without adjusting the proportion of training data nor applying bidirectional learning) does not make any difference from the baseline; also, bidirectional learning must be built upon DTS.

From Columns 1--3, one can learn the importance of the DTS framework. When DTS is not established, simply increasing the proportion of target data (Col~2) incurs an accuracy drop on both GTA$\rightarrow$Cityscapes and SYNTHIA$\rightarrow$Cityscapes. This validates our motivation, \textit{i.e.}, there is a conflict between the \textbf{learning} and \textbf{adapting} abilities. However, after the dual teacher-student framework is established (Col~3, where the first teacher-student group focuses on \textbf{learning} and the second group focuses on \textbf{adapting}), the conflict is weakened and the accuracy gain becomes clear ($1.6\%$ on GTA$\rightarrow$Cityscapes and $2.4\%$ on SYNTHIA$\rightarrow$Cityscapes).

From Columns 3--5, one can observe the contribution of \textit{`Focus on Your Target'} (\textit{i.e.}, increasing the proportion of target data) and bidirectional learning on top of the DTS framework. The best practice is obtained when all the modules are used (Col~5), aligning with the results reported in Tables~\ref{tab:GTA} and~\ref{tab:SYNTHIA}. Interestingly, when bidirectional learning is adopted alone (Col~4), the accuracy gain is much smaller than that of using \textit{`Focus on Your Target'} alone, because the two groups are using the same proportion of data mixing, and thus the bidirectional learning carries limited complementary information.

\begin{table}[!t]
\centering
\begin{tabular}{cc|ccc}
\toprule
\multicolumn{2}{c|}{Baseline}                         & DAFormer      & HRDA          & MIC           \\
\midrule
\multicolumn{1}{c|}{\multirow{2}{*}{GTAv}}    & \textit{Setting A}  & 70.3          & 74.7          & \textbf{76.5}       \\
\multicolumn{1}{c|}{}                         & \textit{Setting B}  & \textbf{71.2} & \textbf{75.2} & 76.2  \\
\midrule
\multicolumn{1}{c|}{\multirow{2}{*}{SYNTHIA}} & \textit{Setting A}  & \textbf{63.8} & \textbf{66.5} & \textbf{67.8}       \\
\multicolumn{1}{c|}{}                         & \textit{Setting B} & 61.2          & 66.0          & 67.2   \\
\bottomrule
\end{tabular}
\caption{The results of DTS with different data mixing options. For SYNTHIA$\rightarrow$Cityscapes, the mIoUs are computed over $16$ classes.}
\label{tab:options}
\end{table}

\begin{figure}[!t]
\centering
\includegraphics[width=0.48\textwidth]{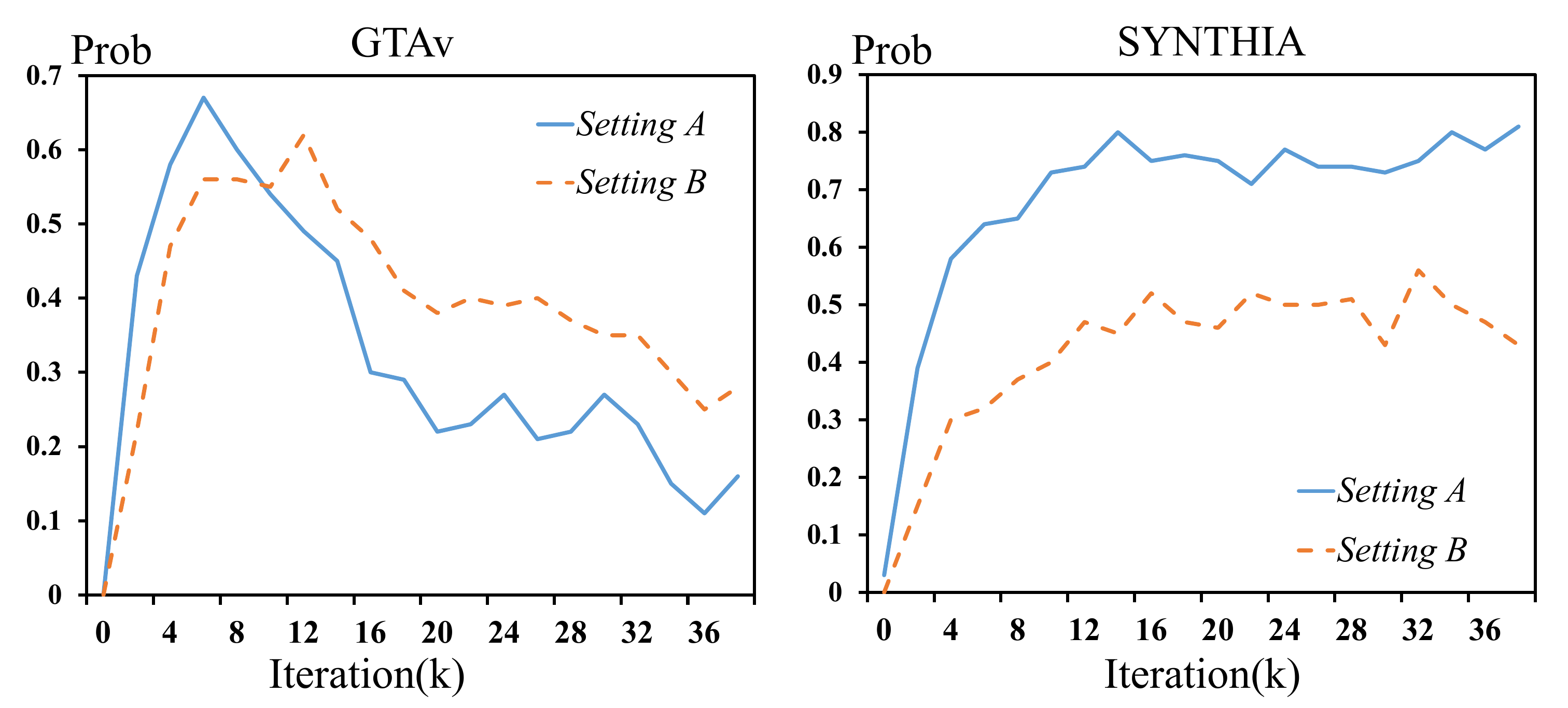}
\caption{The curves of $\mathrm{Prob}$  computed on the GTAv$\rightarrow$Cityscapes and SYNTHIA$\rightarrow$Cityscapes benchmarks. DAFormer is used as the baseline. These curves suggest that, on DAFormer, \textit{Setting B} is better for GTAv$\rightarrow$Cityscapes while \textit{Setting A} is better for SYNTHIA$\rightarrow$Cityscapes.}
\label{fig:options}
\end{figure}

\noindent\textbf{Choose the option for data combination.}\quad
We validate that the data combination option can be chosen by comparing the pseudo-label confidence, as described in Section~\ref{approach:details}. We plot the $\mathrm{Prob}$ curves throughout training procedure of \textit{Setting A} and \textit{Setting B} in Figure~\ref{fig:options}. On GTAv$\rightarrow$Cityscapes, the $\mathrm{Prob}$ value of \textit{Setting B} is eventually higher than that of \textit{Setting A}, and DTS with \textit{Setting B} outperforms DTS with \textit{Setting A} by a $0.9\%$ mIoU (see Table~\ref{tab:options}). However, on SYNTHIA$\rightarrow$Cityscapes, the $\mathrm{Prob}$ value of \textit{Setting A} is always higher than that of \textit{Setting B}, and DTS with \textit{Setting A} outperforms DTS with \textit{Setting B} by a $2.4\%$ mIoU. Table~\ref{tab:options} also shows the segmentation accuracy on the other two baselines. All choices of advantageous results align with the aforementioned criterion.

\begin{table}[!t]
\centering
\begin{tabular}{cc|cc|c}
\toprule   
\multicolumn{2}{c|}{Group \#1} & \multicolumn{2}{c|}{Group \#2} & \multirow{2}{*}{mIoU} \\
$f_{2,t}^\mathrm{st}(\cdot)$ & $f_{2,t}^\mathrm{te}(\cdot)$ & $f_{1,t+1}^\mathrm{st}(\cdot)$ & $f_{1,t+1}^\mathrm{te}(\cdot)$ & \\
\midrule
& \checkmark & & \checkmark & \textit{68.5}$^\dagger$ \\
\checkmark & & & \checkmark & 69.0 \\
& \checkmark & & \checkmark & 69.5 \\
\checkmark & & \checkmark & & 70.2 \\
\rowcolor{Gray}
& \checkmark & \checkmark & & \textbf{71.3} \\
\bottomrule
\end{tabular}
\caption{UDA segmentation accuracy (mIoU, \%) using different models to generate pseudo labels on GTAv$\rightarrow$Citysacpes. $^\dagger$: In the first row, each group only uses the teacher model from the other group; in other cases, each group also uses its own teacher models (not presented in the table).}
\label{tab:pseudolabel}
\end{table}

\noindent\textbf{Models for pseudo label generation.}\quad
We study the impact of different models for generating pseudo labels. Results are summarized in Table~\ref{tab:pseudolabel}. When each teacher-student group only uses the teacher model from another group (the first row), there is almost no accuracy gain (the baseline is $68.3\%$), indicating the necessity of the supervision signals from its own teacher.

Then, we discuss the situations where each group uses its teacher and one model from the other group. For the second teacher-student group, using $f_{1,t+1}^\mathrm{st}(\cdot)$ as the teacher is superior to using $f_{1,t+1}^\mathrm{te}(\cdot)$, because the former model is optimized one more iteration and thus produces better semantics (\textit{i.e.}, `learning from future'). For the first teacher-student group, only models from the current iteration of the second group can be used, and $f_{2,t}^\mathrm{te}(\cdot)$ always works better than $f_{2,t}^\mathrm{st}(\cdot)$, implying the advantage of EMA.

\begin{table}[!t]
\centering
\begin{tabular}{l|cc|cc}
\toprule
\multirow{2}*{Method} & \multicolumn{2}{c|}{GTAv$\rightarrow$City.} & \multicolumn{2}{c}{SYNTHIA$\rightarrow$City.} \\
{} & DACS & DAFormer & DACS & DAFormer \\
\midrule
w/o DTS & 53.9 & 56.0 & 54.2 & 53.5 \\
\rowcolor{Gray}
w DTS & 55.2 & 59.9 & 56.6 & 56.9 \\
\bottomrule
\end{tabular}
\caption{UDA segmentation accuracy (mIoU, \%) on two CNN-based backbones and two benchmarks.}
\label{tab:CNN}
\end{table}

\noindent\textbf{DTS applied to other backbones.}\quad
To show that DTS adapts to different network backbones (\textit{e.g.}, CNNs), we follow the previous approaches~\cite{yang2020fda,tranheden2021dacs,huo2022domain} to inherit a ResNet101~\cite{he2016deep} pre-trained on ImageNet and MSCOCO and train it using DeepLabV2~\cite{chen2017deeplab}. The other settings remain unchanged. Results are listed in Table~\ref{tab:CNN}. Still, DTS brings consistent accuracy gains, improving DACS and DAfromer by $1.3\%$ and $3.9\%$ on GTAv$\rightarrow$Cityscapes, and by $2.4\%$ and $3.4\%$ on SYNTHIA$\rightarrow$Cityscapes. Class-wise comparison can refer to 
Appendix~\ref{appendix: CNN results}. These experiments validate that DTS is a generalized UDA framework that does not rely on specific network architectures.

\noindent\textbf{Application to semi-supervised semantic segmentation.}\quad
As a generalized self-training framework, DTS can also be applied to semi-supervised segmentation. We conduct experiments on PASCAL VOC~\cite{everingham2015pascal} and follow the conventional settings to train a DeepLabV3+ model with ResNet101 as the backbone. Other settings remain unchanged; please refer to Appendix~\ref{appendix: semi details} for details. We report the segmentation results in Table~\ref{tab:semi-voc}. Using $1/16$, $1/8$, $1/4$ labeled data, DTS produces consistent accuracy gains of $3.5\%$, $1.7\%$, $0.8\%$ over the MeanTeacher approach~\cite{tarvainen2017mean}, and $1.5\%$, $1.3\%$, $1.0\%$ over the re-implemented FST baseline~\cite{du2022learning}. This validates that semi-supervised learning also benefits from bidirectional learning which effectively improves the ability to learn from the pseudo labels generated from other groups.


\begin{table}[!t]
\centering
\setlength{\tabcolsep}{0.12cm}
\begin{tabular}{l|ccc}
\toprule
Methods      & 1/16 (662)        & 1/8 (1323)           & 1/4 (2646)           \\
\midrule
Mean Teacher~\cite{tarvainen2017mean} & 71.3          & 75.4          & 76.5          \\
FST~\cite{du2022learning}          & 73.9          & 76.1          & \textbf{78.1} \\
\midrule
FST*         & 73.3          & 75.8          & 76.3          \\
\rowcolor{Gray}
DTS (ours)    & \textbf{74.8} & \textbf{77.1} & 77.3          \\
\bottomrule
\end{tabular}
\caption{Semi-supervised segmentation mIoU (\%) on PASCAL VOC.  *: This is our re-implementation of FST.}
\label{tab:semi-voc}
\end{table}

\section{Conclusions}

In this paper, we study the UDA segmentation problem from the perspective of endowing the model with two-fold abilities, \textit{i.e.}, \textbf{learning} from the reliable semantics, and \textbf{adapting} to the target domain. We show that these two abilities can conflict with each other in a single teacher-student model, and hence we propose a dual teacher-student (DTS) framework, where the two models stick to the `learning ability' and `adapting ability', respectively. Specifically, the second model learns to `focus on your target' to improve its accuracy in the target domain. We further design a bidirectional learning strategy to facilitate the two models to communicate via generating pseudo labels. We validate the effectiveness of DTS and set new records on two standard UDA segmentation benchmarks.

As the accuracy on the two benchmarks fast approaches the supervised baseline (\textit{e.g.}, a $77.8\%$ mIoU on Cityscapes), we look forward to building larger and higher-quality synthesized datasets to break through the baseline. This is of value to real-world applications, as synthesized data can be generated at low costs. We expect that the proposed methodology can be applied to other scenarios such as domain generalization, with the second group focusing on a generic, unbiased feature space.


{\small
\bibliographystyle{ieee_fullname}
\bibliography{egbib}
}

\appendix


\begin{table*}[!t]
\centering
\resizebox{\textwidth}{!}{
\begin{tabular}{lcccccccccccccccccccc}
\toprule
\multicolumn{1}{l|}{\textbf{Method}} &
  \rotatebox{90}{Road}&
  \rotatebox{90}{Sidewalk} &
  \rotatebox{90}{Building} &
  \rotatebox{90}{Wall} &
 \rotatebox{90}{Fence} &
 \rotatebox{90}{Pole} &
  \rotatebox{90}{Light} &
  \rotatebox{90}{Sign} &
  \rotatebox{90}{Veg} &
  \rotatebox{90}{Terrain} &
 \rotatebox{90}{ Sky }&
  \rotatebox{90}{Person }&
  \rotatebox{90}{Rider} &
  \rotatebox{90}{Car }&
  \rotatebox{90}{Truck} &
  \rotatebox{90}{Bus} &
  \rotatebox{90}{Train} &
  \rotatebox{90}{Motor} &
  \multicolumn{1}{c|}{\rotatebox{90}{Bike}} &
  \textbf{mIOU} \\ \hline \hline
  
\multicolumn{1}{l|}{DACS}            &  
95.3  &  	67.9  &  	87.5  &  	33.7  &  	30.5  &  	40.2  &  	50.2 	 &  56.4  &  	87.7 &   	45.3  &  	87.0  &  	67.4  &  	29.7  &  	89.5 &   	48.0  &  	50.5  &  	2.2  &  	23.0  &  	\multicolumn{1}{c|} {32.0}  &  	53.9 
\\

\rowcolor{Gray}\multicolumn{1}{l|}{DACS + DTS}            & 96.1  &	72.2  &	88.2  &	38.1  &	33.6  &	41.6  &	52.1 	 &61.4  &	88.6  &	50.0  & 89.1  &	68.5  & 38.0  &	90.6  &	59.1  &	58.0  &	0.2  &	8.6 	&\multicolumn{1}{c|} {14.5} 	 &\textbf{55.2}  \\  

\multicolumn{1}{l|}{DAFormer}            & 
94.7  & 	66.3 	 & 87.9  & 	40.7  & 	33.9 	 & 37.2 	 & 50.2  & 	52.9 	 & 87.9 	 & 46.5 	 & 88.2  & 	69.8 	 & 44.2  & 	89.1 	 & 43.1 	 & 55.8  & 	0.7  & 	24.7  & 	\multicolumn{1}{c|} {50.7}  &  56.0 \\

\rowcolor{Gray}\multicolumn{1}{l|}{DAFormer+DTS}            & 96.4  &		73.9  &		88.6 &	40.0 &	39.8 &	42.2 &	52.2& 	63.5 &	88.8 &	49.6 &	89.3 	& 70.6 &	45.6 &	90.8 &	61.1 	& 57.0 	 &	0.4 &	 	33.1 	&\multicolumn{1}{c|} {55.4}  &		\textbf{59.9} 
 \\ \midrule
 
\multicolumn{1}{l|}{DACS}            & 
 81.3  & 	38.9  & 	84.6 &  	15.3 	 & 1.7 	 & 40.2 &  	45.2  & 	50.3  & 	85.0 & --  & 	85.3  & 	70.4  & 	41.9 	 & 84.6 & --  & 	44.8 & -- 	 & 39.8  & 	\multicolumn{1}{c|} {57.5 } & 	54.2 \\

\rowcolor{Gray}\multicolumn{1}{l|}{DACS + DTS}            & 88.7  & 	52.7  & 85.5    & 7.2  & 2.5  & 40.5  & 48.9  & 52.1 & 86.0 &-- & 87.8    & 72.5  & 46.8  & 83.9 & --  & 43.4 &-- & 46.6  & \multicolumn{1}{c|}{60.9 }
& \textbf{56.6}
\\  

\multicolumn{1}{l|}{DAFormer}            & 66.8  & 	29.3  & 	85.0  & 	19.1  & 	2.3  & 	38.7 	 & 45.9 	 & 51.6  & 	80.8 &-- 	 & 85.9 	 & 70.2 	 & 41.9  &   	84.9 &--  & 	46.0 &-- 	 & 48.9 	 & \multicolumn{1}{c|}{58.4 } & 	53.5 \\

\rowcolor{Gray} \multicolumn{1}{l|}{DAFormer+DTS}            &  88.9  & 	52.5  & 85.1  & 7.5 & 2.4  & 39.7  & 49.5 & 52.7  & 85.6 &--  & 87.0  & 72.8  & 47.0  & 85.0 &--  & 48.0 &--   & 47.7  & \multicolumn{1}{c|}{58.9} & \textbf{56.9 }
\\
 
\bottomrule
\end{tabular}}
\caption{Segmentation accuracy (IOU, \%) of baselines~\cite{tranheden2021dacs,hoyer2022daformer} and DTS based on ResNet101 backbone. The top part shows the transfer results for \textbf{GTAv$\rightarrow$Cityscapes} and the bottom part shows the results for \textbf{SYNTHIA$\rightarrow$Cityscapes}, where mIoU is computed over 16 classes. All results are averaged over $3$ runs.}
\label{tab:CNN-comparsion}
\end{table*}

\section{Examples of the Conflict in Learning}
\label{appendix: conflict}

\begin{figure}[!t]
\centering
\includegraphics[width=0.48\textwidth]{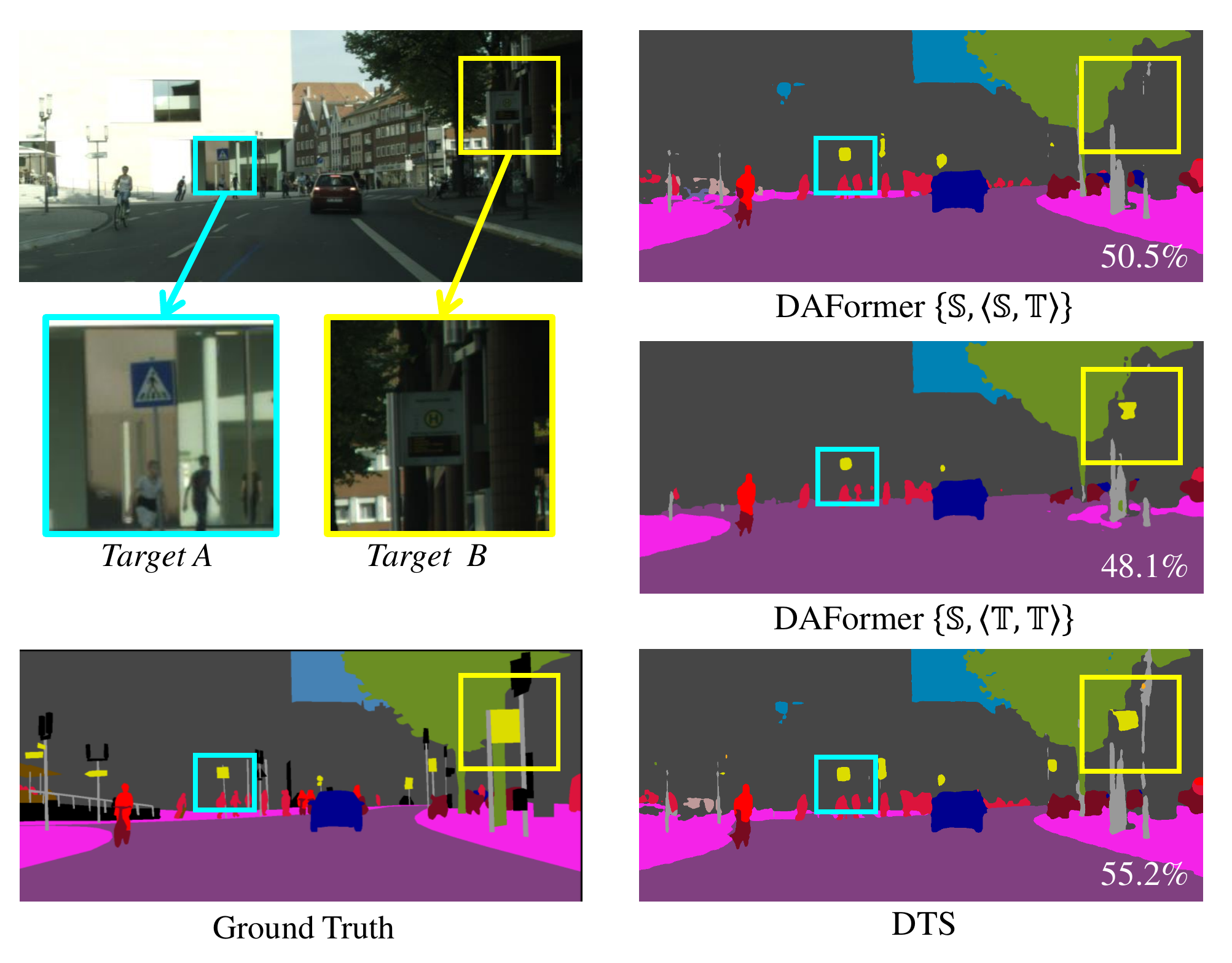}
\caption{Qualitative and quantitative comparisons of the single teacher-student model trained on different data combination strategies. The numbers shown in the bottom-right corner denote the mIoU of the overall image.}
\label{fig:conflict-supply}
\end{figure}

We provide an example of the conflict between the learning and adapting abilities, which we discussed in Section~\ref{sec: introduction} of the main article. This is tested using a single teacher-student framework. We use the same test case as in Figure~\ref{fig: motivation}, which has two targets of \textsf{traffic sign} with different appearances. When trained on the original data combination, $\{\mathbb{S},\langle\mathbb{S},\mathbb{T}\rangle\}$, the baseline method (DAFormer~\cite{hoyer2022daformer}) shows that \textit{Target B} (an object with a larger transfer difficulty) gradually grows but eventually drops to $0$, resulting in an overall mIoU of $50.5\%$. When the proportion of the target domain is increased using the combination of $\{\mathbb{S},\langle\mathbb{T},\mathbb{T}\rangle\}$, the model has a better performance on \textit{Target B} but reports a lower mIoU of $48.1\%$, as shown in Figure~\ref{fig:conflict-supply}. Therefore, improving the adaptability of \textit{Target B} in this framework can potentially deteriorate the learning of other semantic concepts. However, the proposed DTS performs well on \textit{Target B} without harming the recognition of other classes, hence improving the overall mIoU to $55.2\%$.

\section{Statistical Differences between Domains}
\label{appendix: class frequency}
\begin{figure}[!t]
\centering
\includegraphics[width=0.42\textwidth]{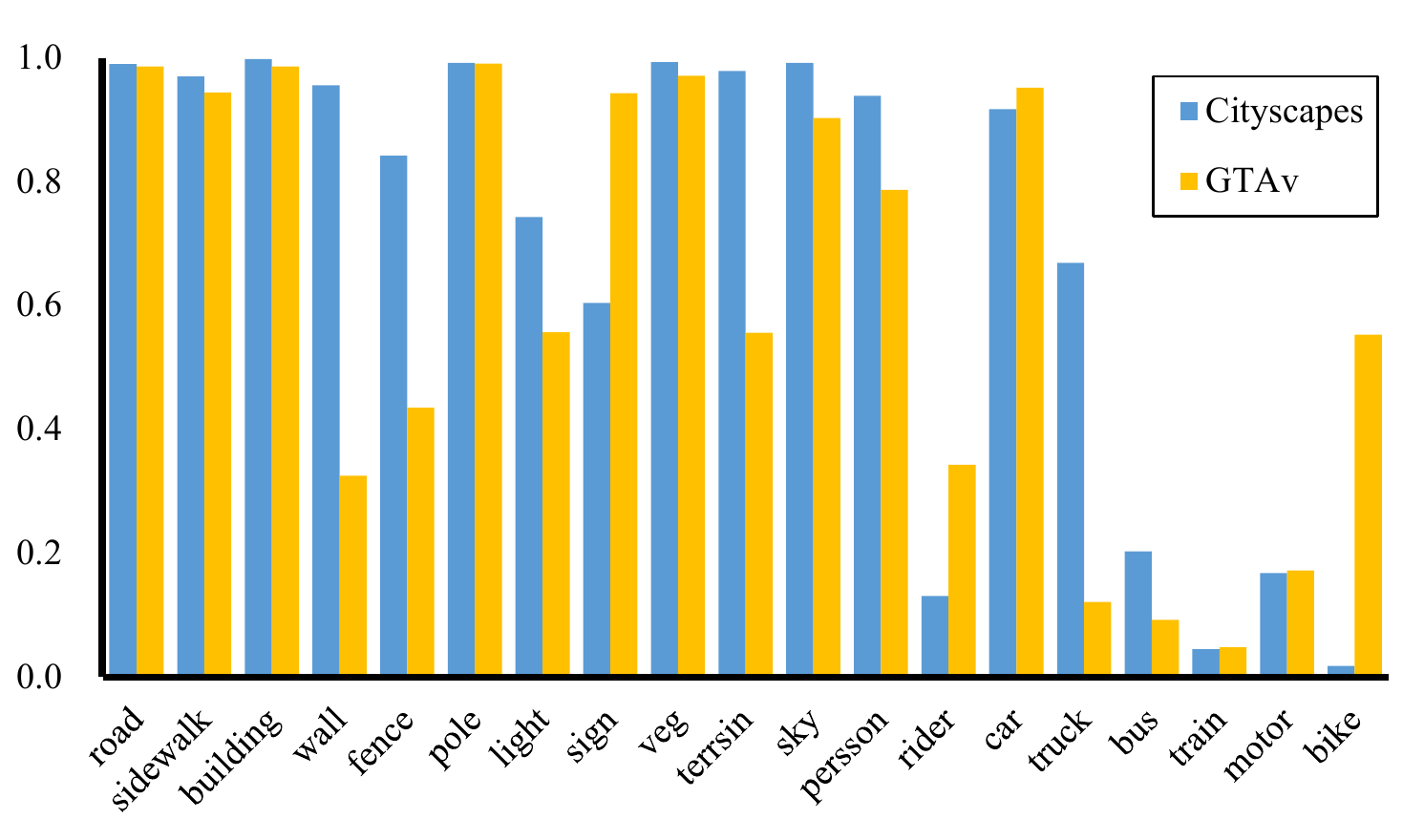}
\caption{Frequency of subclass occurrence in the source dataset (GTAv) and target dataset (Cityscapes).}
\label{fig:class-frequency}
\end{figure}

To make a clear and distinct comparison between the source and target domains, we compute the occurrence frequency of each class in GTAv and Cityscapes. This is done by simply dividing the number of images including a given category by the total number of images -- we believe that better metrics can be defined. The comparison of $19$ classes is presented in Figure~\ref{fig:class-frequency}. We can see that the \textsf{traffic sign} and \textsf{truck} classes exhibit a large difference between the two datasets, which is consistent with the difficulties during transfer (see the analysis in Section~\ref{sec: comparison with sota} of the main article). Except for these two representative classes, we also find that the \textsf{fence}, \textsf{traffic light}, \textsf{rider}, and \textsf{bike} classes are statistically improved by DTS on three strong baselines (see Table~\ref{tab:GTA}), despite them having substantial differences between GTAv and Cityscapes.

\section{Results on the CNN-based Backbones}
\label{appendix: CNN results}
We present the class-wise segmentation accuracy of the CNN-based backbones on both the GTAv$\rightarrow$Cityscapes and SYNTHIA$\rightarrow$Cityscapes benchmarks in Table~\ref{tab:CNN-comparsion}. The proposed DTS approach achieves significant improvements in the \textsf{traffic sign} and \textsf{truck} categories when GTAv is used as the source data, which is consistent with the results obtained using a transformer backbone (see the analysis in the previous section). In the case of using SYNTHIA as the source data, our approach outperforms other methods in the \textsf{road}, \textsf{sidewalk}, \textsf{person}, and \textsf{rider} classes, which are easily confused with each other (\textsf{road} \textit{vs.} \textsf{sidewalk}, \textsf{person} \textit{vs.} \textsf{rider}).

\begin{table}[!t]
\centering
\begin{tabular}{c|c|c}
\toprule
Options                          &  GTAv         & SYNTHIA      \\ \hline\hline
$\{\langle\mathbb{T},\mathbb{T}\rangle\}$   & 69.1          & 61.1          \\
$\{\langle\mathbb{S},\mathbb{T}\rangle\}$   & 69.3          & 59.4          \\
$\{\mathbb{S},\langle\mathbb{T},\mathbb{T}\rangle\}$                                 & 70.3          & \textbf{63.8} \\
$\{\langle\mathbb{S},\mathbb{T}\rangle,\langle\mathbb{T},\mathbb{T}\rangle\}$                            & \textbf{71.2} & 61.2          \\ \bottomrule
\end{tabular}
\caption{The results of DTS with more data mixing options. We show the mIoU of the $16$ classes in SYNTHIA and make this comparison based on the DAFormer baseline. }
\label{tab:more-data-mixings}
\end{table}

\section{More Data Combination Options}
\label{appendix: more data mixing}
We propose two data combinations to tune the proportion of target data in our method: $\{\mathbb{S},\langle\mathbb{T},\mathbb{T}\rangle\}$ and $\{\langle\mathbb{S},\mathbb{T}\rangle,\langle\mathbb{T},\mathbb{T}\rangle\}$. While there are two other options to increase the focus on the target domain as well, namely $\{\langle\mathbb{T},\mathbb{T}\rangle\}$  and $\{\langle\mathbb{S},\mathbb{T}\rangle\}$. The two options only include one type of data and have a disadvantage compared to the previous two combinations (see Table~\ref{tab:more-data-mixings}). $\{\langle\mathbb{T},\mathbb{T}\rangle\}$  means that the second teacher-student group achieves the ability of \textbf{learning} only from the pseudo labels provided by $f_1^\mathrm{st}(\mathbf{x};\boldsymbol{\theta}_1^\mathrm{st})$. As a result, noise can easily persist without accurate supervision from the source data.  $\{\langle\mathbb{S},\mathbb{T}\rangle\}$ has a similar data proportion as $\{\mathbb{S},\langle\mathbb{T},\mathbb{T}\rangle\}$ but cannot solve the problem of inconsistency between the training and testing data. Using $\langle\mathbb{T},\mathbb{T}\rangle$ as the training domain is an effective way to bridge the gap between the training and testing phases. Therefore, the two combinations containing two different types of data used in DTS are more suitable for balancing the \textbf{learning} and \textbf{adapting} abilities.

\section{Details of Semi-supervised Segmentation}
\label{appendix: semi details}
\noindent
\textbf{Dataset.} We perform semi-supervised segmentation on the widely used PASCAL VOC dataset~\cite{everingham2015pascal}, which comprises $20$ object classes and a background class. The training, validation, and testing sets contain $1\rm{,}464$, $1\rm{,}449$, and $1\rm{,}456$ images, respectively. To align with prior works, we use the augmented set, which contains $10\rm{,}582$ images, as the complete training set. During the training, we apply random flipping and cropping to augment the image data and resize the image to $513\times513$ before feeding it to the models.

\noindent
\textbf{Implementation.} We perform all experiments using the DeepLabV3+~\cite{chen2018encoder} network with a ResNet-101 backbone pre-trained on ImageNet. An SGD optimizer, with a learning rate of $0.0001$ for the encoder and $0.001$ for the decoder, and a weight decay of $0.0001$, is used. The batch size of both labeled and unlabeled data is $16$ for each group teacher-student. We train the models for $40$ epochs and report the average of three runs for all the results. Since there is no domain gap between the labeled and unlabeled parts, only a dual teacher-student framework and bidirectional learning are applied in the semi-supervised experiments.

\end{document}